\documentclass[a4paper]{article}

\usepackage{multirow}
\usepackage{bm}
\usepackage{mathrsfs}
\usepackage{caption}
\usepackage{algorithm}
\usepackage{algorithmic}

\usepackage{INTERSPEECH2021}

\title{Discriminative Self-training for Punctuation Prediction}
\name{Qian Chen, Wen Wang, Mengzhe Chen, Qinglin Zhang}
\address{Speech Lab, Alibaba Group}
\email{\{tanqing.cq, w.wang, mengzhe.cmz, qinglin.zql\}@alibaba-inc.com}

\begin{document}

\maketitle
\begin{abstract}
Punctuation prediction for automatic speech recognition (ASR) output transcripts plays a crucial role for improving the readability of the ASR transcripts and for improving the performance of downstream natural language processing applications. However, achieving good performance on punctuation prediction often requires large amounts of labeled speech transcripts, which is expensive and laborious. In this paper, we propose a Discriminative Self-Training approach with weighted loss and discriminative label smoothing to exploit unlabeled speech transcripts. Experimental results on the English IWSLT2011 benchmark test set and an internal Chinese spoken language dataset demonstrate that the proposed approach achieves significant improvement on punctuation prediction accuracy over strong baselines including BERT, RoBERTa, and ELECTRA models.  The proposed Discriminative Self-Training approach outperforms the vanilla self-training approach. We establish a new state-of-the-art (SOTA) on the IWSLT2011 test set, outperforming the current SOTA model by 1.3\% absolute gain on F$_1$.

\end{abstract}

\section{Introduction}
\label{sec:intro}
Spoken language transcripts generated by automatic speech recognition (ASR) systems usually have no punctuation marks. However, many downstream applications, such as machine translation and dialogue systems, are usually trained on well-formed text with proper punctuation marks. Hence, there is a significant mismatch between the training corpora and the actual ASR transcript input to these applications, causing dramatic performance degradation. In addition, lack of punctuation marks reduces the readability of speech transcripts. Consequently, punctuation prediction has become a crucial post-processing task for speech transcripts.

A critical challenge for punctuation prediction is that to achieve good performance on actual ASR output, many existing approaches require large amounts of human-annotated spoken language data. However, acquiring labeled spoken language data is a costly process. 
There have been several approaches to alleviate this issue through transfer learning, such as using large amounts of written language data with punctuation marks for self-supervised training for punctuation prediction~\cite{DBLP:conf/icassp/ChenCLW20,DBLP:conf/apsipa/MakhijaHC19,DBLP:journals/corr/abs-2004-00248}. However, there exists significant discrepancy between written language data and spoken language data. A large amount of spoken language data is available for training ASR models; yet many spoken language datasets, such as LibriSpeech \cite{DBLP:conf/icassp/PanayotovCPK15}, do not have transcripts with manually labeled punctuation marks.  In this work, we propose a Discriminative Self-Training (denoted Disc-ST) approach for exploiting unlabeled speech transcripts without punctuation marks and demonstrate its effectiveness in improving strong baselines for punctuation prediction. 

Our contributions can be summarized as follows: 

1) To the best of our knowledge, this paper is the first to explore self-training for punctuation prediction. We exploit large-scale unlabeled spoken language data without punctuation, such as transcripts used for training ASR systems, through self-training to improve strong baseline models based on BERT, RoBERTa, and ELECTRA. 

2) We propose a Discriminative Self-Training approach using weighted loss and discriminative label smoothing when training on combined human-labeled and pseudo-labeled data. 

3) Experimental results on the English IWSLT2011 benchmark test set and an internal Chinese spoken language dataset demonstrate that our approach achieves significant improvement on punctuation prediction accuracy over strong baselines including BERT, RoBERTa, and ELECTRA models.  The proposed discriminative self-training approach outperforms the vanilla self-training approach. We establish a new state-of-the-art (SOTA) on the IWSLT2011 test set, outperforming the current SOTA model by 1.3\% absolute on F$_1$. 

\section{Related Work}
\label{sec:related}
Previous punctuation prediction models can be categorized into three major categories. The first category views punctuation prediction as hidden inter-word event detection, using models such as n-gram language models~\cite{DBLP:conf/icassp/BeefermanBL98} and Hidden Markov Models (HMMs) \cite{christensen2001punctuation,DBLP:journals/taslp/LiuSSHOH06}. The second category treats punctuation prediction as sequence labeling by assigning a punctuation mark to each word using conditional random fields (CRFs) \cite{DBLP:journals/taslp/LiuSSHOH06,DBLP:conf/emnlp/LuN10,DBLP:conf/interspeech/UeffingBV13}, convolutional neural networks (CNNs)~\cite{DBLP:conf/lrec/CheWYM16,DBLP:conf/interspeech/ZelaskoSMSCD18}, recurrent neural networks (RNNs)~\cite{DBLP:conf/interspeech/ZelaskoSMSCD18,DBLP:conf/interspeech/TilkA15} and their variants \cite{DBLP:conf/interspeech/TilkA16,DBLP:conf/interspeech/YiTWL17}. The third category uses sequence-to-sequence modeling in which the source is unpunctuated text and the target is punctuated text \cite{DBLP:conf/iwslt/PeitzFMN11} or sequences of punctuation marks \cite{DBLP:conf/icassp/YiT19,DBLP:conf/slt/Klejch0R16}.  

Some previous punctuation prediction approaches explored lexical information and acoustic/prosodic information separately and in combination.  An important observation from these works is that information from text plays a much more crucial role in punctuation prediction than prosodic information; still, adding prosodic information to text-based modeling may achieve performance improvement~\cite{DBLP:conf/interspeech/HuangZ02,DBLP:journals/csl/LiuCHSS06}. In this work, we focus on improving text-based punctuation prediction models over the current SOTA which only explores text information. 

Self-supervised learning can help reduce the required amount of labeling. A model pre-trained on unlabeled data through self-supervised learning can be fine-tuned on smaller amount of labeled data. Recently there has been much research exploring this pretraining-finetuning framework, built upon transformer-based pre-trained language models~\cite{DBLP:conf/apsipa/MakhijaHC19,DBLP:conf/interspeech/YiTTBF20,DBLP:journals/corr/abs-2004-00248,DBLP:conf/aclnut/AlamKA20, DBLP:conf/iwslt/CourtlandFM20}. 
Different from these previous studies, our work explores self-training(ST)~\cite{DBLP:conf/acl/Yarowsky95} in finetuning transformer-based pre-trained models for punctuation prediction. 

In the standard procedure of ST, a teacher model trained with labeled data is used to generate pseudo labels on unlabeled data. Then the data with pseudo labels are combined with labeled data to train a student model. 
ST has proven effective on many tasks, such as image classification \cite{DBLP:conf/cvpr/XieLHL20}, word sense disambiguation \cite{DBLP:conf/acl/Yarowsky95}, disfluency detection \cite{Wang2020}, parsing \cite{DBLP:conf/naacl/McCloskyCJ06}, machine translation \cite{DBLP:conf/iclr/HeGSR20}, and text summarization \cite{DBLP:conf/iclr/HeGSR20}. 
Different from the tasks on which ST was applied in previous works, this work investigates the efficacy of applying ST on punctuation prediction. On top of the vanilla ST approach, we explore discriminative weighted loss and labeling smoothing to improve the effectiveness of self-training. 

\section{Proposed Approach}
\label{sec:approach}
\subsection{Model Architecture}

Our model treats punctuation prediction as a sequence labeling task. The input are transcripts without punctuation. For languages without explicit word boundaries such as Chinese, the input is segmented into words. Our model predicts whether there is a specific punctuation mark after each word. The word tokens are processed with WordPiece tokenization and the output are fed into a Transformer encoder~\cite{DBLP:conf/nips/VaswaniSPUJGKP17}. The final hidden state of the encoder corresponding to the last sub-token is used as the input to the softmax classifier for punctuation prediction.

\subsection{Discriminative Self-Training}
\label{subsec:architecture}
\begin{figure}[t]
\centering
\includegraphics[width=0.45\textwidth]{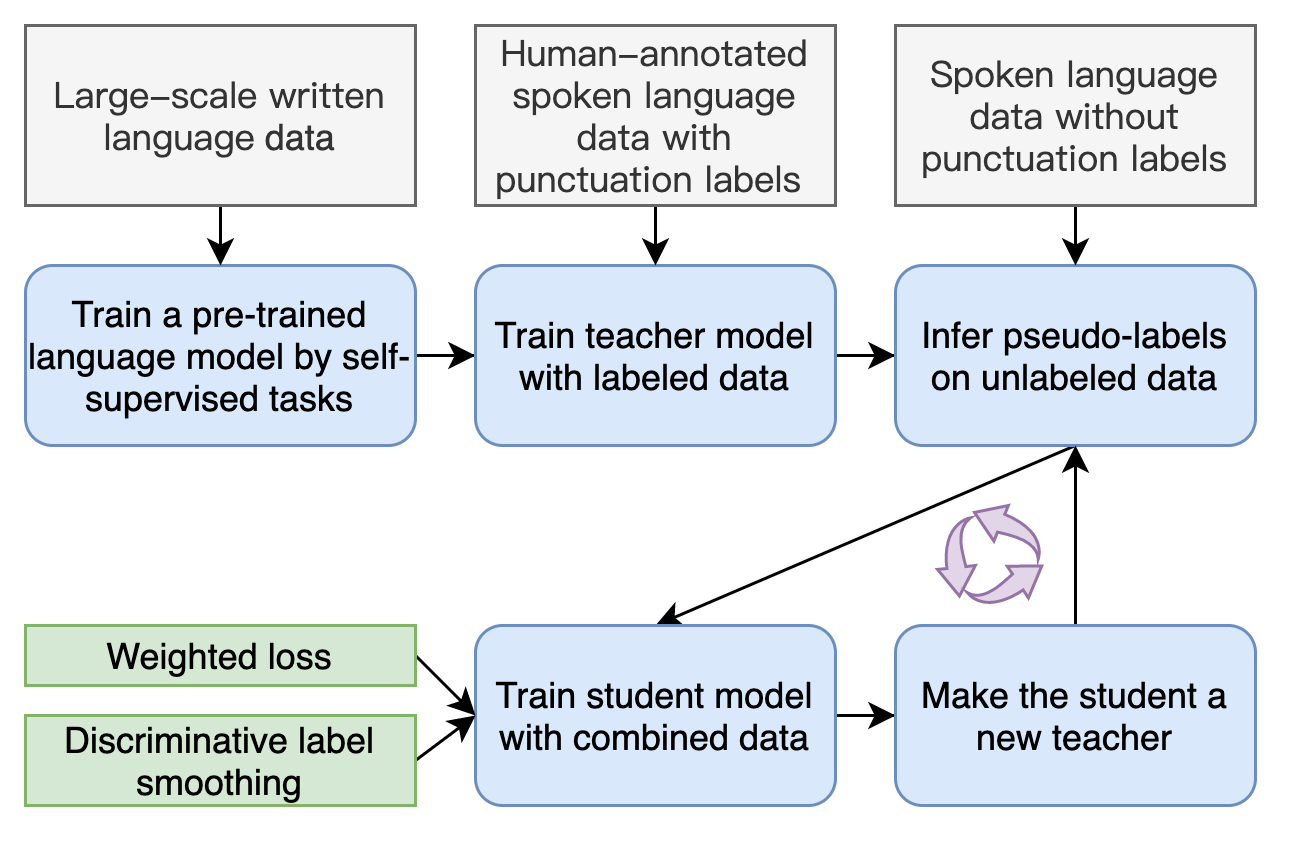}
\caption{Illustration of the proposed discriminative self-training approach for punctuation prediction.}
\label{fig:noisyselftraining}
\end{figure}

Figure~\ref{fig:noisyselftraining} provides an overview of the proposed discriminative self-training approach for punctuation prediction, 
given large-scale written language data, human-labeled spoken language data with punctuation labels ${(x_1, y_1), (x_2, y_2),\dots,(x_n,y_n)}$ and unlabeled spoken language data ${\tilde{x}_1, \tilde{x}_2,\dots, \tilde{x}_m}$.

First, we pre-train a language model on large-scale well-formed text corpora by self-supervised tasks, such as masked language modeling (MLM) and next sentence prediction (NSP), to obtain deep bidirectional language representations. 

We then initialize a teacher model $\theta^t$ from the pre-trained language model and train it on spoken language data with human-labeled punctuation marks through minimizing the cross-entropy loss $\ell$:
\begin{align}
\ell = \frac{1}{n}\sum_{i=1}^n{\ell(y_i, f(x_i, \theta^t))}
\end{align}

\noindent where $f$ denotes the classifier.
Next, we infer pseudo labels on clean unlabeled spoken language data as follows: 
\begin{align}
\tilde{y}_i = f(\tilde{x}_i, \theta^t)), \forall i=1,\dots,m
\end{align}

Then, we initialize a student model $\theta^s$ from pre-trained language models and train it by minimizing the cross-entropy loss on the combination of human-labeled spoken language data and pseudo-labeled spoken language data.

In this paper, we propose a self-training approach employing weighted loss and label smoothing in a discriminative way when training on the combination of human-labeled data and pseudo-labeled data, denoted discriminative self-training (\textbf{Disc-ST}). 
Intuitively, the pseudo-labeled data has more noise than human-labeled data. Thus, we use different weights to combine the loss on human-labeled data and pseudo-labeled data (denoted \textbf{weighted loss}), as follows:
\begin{align}
\ell = \sum_{i=1}^n{\ell(y_i, f(x_i, \theta^s))} + \alpha \sum_{i=1}^m{\ell(\tilde{y}_i, f(\tilde{x}_i, \theta^s))}
\end{align}
where $\alpha$ is a weight factor.
Meanwhile, we explore one of the output regularizers, label smoothing \cite{DBLP:conf/cvpr/SzegedyVISW16}, to deal with training with noisy labels. In  future  work,  we plan to  explore other output regularizers such as focal loss \cite{DBLP:journals/pami/LinGGHD20} and bootstrapping loss \cite{DBLP:journals/corr/ReedLASER14}. The standard cross-entropy loss is calculated as 
\begin{align}
\ell &= -\sum_{i=1}^K y_i log(p_i)=-y^Tlog(p)
\end{align}
where $y = \text{One-hot}(i)$, $p = [p_1, p_2, \dots, p_K]$, $K$ is the number of classes,
and $i$ is the label index. For label smoothing, we replace $y$ with $y_{LS}$ as 
\begin{align}
y_{LS} &= (1-\beta) y + \beta/K 
\end{align}

Similar to the motivation for weighted loss, considering the different noise levels in human-labeled data and pseudo-labeled data, we use label smoothing discriminatively, i.e., we use different factors ($\beta_1$ and $\beta_2$) for human-labeled data and pseudo-labeled data. $\alpha$, $\beta_1$ and $\beta_2$ are hyperparameters optimized on the validation set.
After Disc-ST, we put the student back as the teacher, and iterate this process until the performance converges. The student model achieving the best performance on the human-labeled validation set is chosen as our final model.

\subsection{Double-Overlap Sliding Window Decoding Strategy}
\label{subsec:decoding}
\begin{figure*}[t]
\centering
\includegraphics[width=0.85\textwidth]{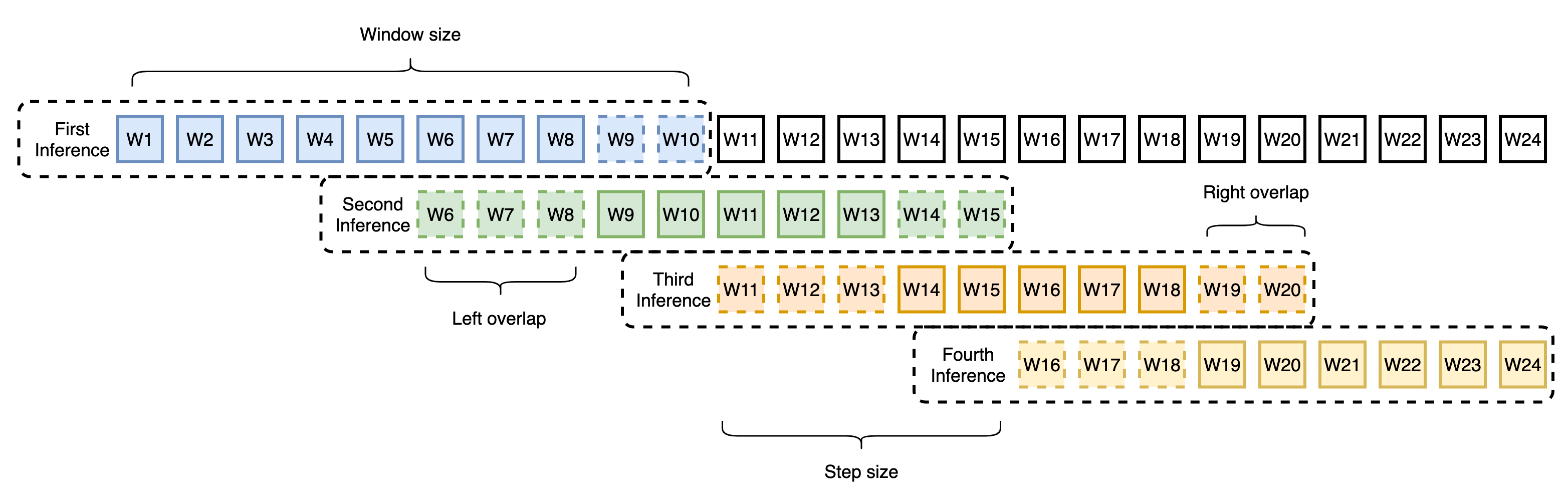}
\caption{Illustration of double-overlap sliding window decoding strategy.}
\label{fig:decoding}
\end{figure*}

When evaluating punctuation prediction performance on a test set, it is infeasible to decode a very long utterance without any segmentation all at once, since it will cause unacceptable latency and out-of-memory issues. Previous work studied different decoding strategies to segment long sequences, including overlapping windows \cite{DBLP:conf/iwslt/ChoNW12}, streaming input scheme \cite{cho2015punctuation}, overlapped-chunk split and merging algorithm \cite{DBLP:journals/corr/abs-1908-02404}, and a fast decoding strategy \cite{DBLP:conf/icassp/ChenCLW20}. We extend the overlapped-chunk split and merging algorithm~\cite{DBLP:journals/corr/abs-1908-02404} to improve punctuation prediction accuracy.
As showed in Figure~\ref{fig:decoding}, we use a sliding window with both left and right overlapped context, and only keep predictions of the model where there is enough context information on both sides. The step size equals the window size minus the sum of the left overlap size and the right overlap size. In our experiments, we observe that the left context is more important than the right context for punctuation prediction accuracy. 
The main difference from the original overlapped-chunk split and merging algorithm is that our step size can be tuned by optimizing the left and right overlap sizes (as two hyperparameters) based on the performance on a validation set; whereas their step size is not tuned and is set as half of the window size.

\section{Experiments}
\subsection{Datasets}
We evaluate the punctuation prediction accuracy on the English IWSLT2011 benchmark dataset and an internal Chinese spoken language dataset.
The IWSLT2011 benchmark contains three types of punctuation marks (comma, period, and question mark). We use the same data organization and same tokenized data as Che et al. \cite{DBLP:conf/lrec/CheWYM16}\footnote{https://github.com/IsaacChanghau/neural\_sequence\_labeling} used.
We also compare the proposed approach with previous methods on a large internal Chinese dataset \cite{DBLP:conf/icassp/ChenCLW20}. We use Jieba\footnote{https://github.com/fxsjy/jieba} for Chinese word segmentation. The punctuation annotations consist of four types of punctuation marks (comma, period, question mark, and enumeration comma).
For unlabeled spoken language data, we use LibriSpeech \cite{DBLP:conf/icassp/PanayotovCPK15}, Fisher Speech Transcripts Part 1 and Part 2~\cite{fisher1and2} for the English dataset and use internal speech transcripts without punctuation for the Chinese dataset.
Data statistics are summarized in Table~\ref{tab:stat}.
We evaluate the punctuation prediction performance using token-based precision (P), recall (R), F$_1$-score (F$_1$), following previous works~\cite{DBLP:conf/interspeech/TilkA15}.
\begin{table}[ht]
\renewcommand{\arraystretch}{1}
\begin{center}
\begin{tabular}{l l r r r}
\hline
\multicolumn{1}{l}{\textbf{Dataset}} & 
\multicolumn{1}{l}{\textbf{Split}} & \multicolumn{1}{l}{\textbf{\#Words}} &
\multicolumn{1}{l}{\textbf{\#Punctuation}}\\
\hline
\multirow{4}{*}{\textbf{IWSLT2011}} & Train & 2M & 301K  \\
& Unlabeled & 30M & -\\
& Dev & 296K & 43K \\
& Test & 13K & 2K \\
\hline
\multirow{4}{*}{\textbf{Chinese}} & Train & 5M & 745K \\
 & Unlabeled & 68M & -   \\
 & Dev &  132K & 18K \\
& Test & 93K & 15K \\
\hline
\end{tabular}
\end{center}
\caption{Statistics of train, unlabeled speech data, dev, and test sets for IWSLT2011 and internal Chinese datasets.}
\label{tab:stat}
\end{table}

\vspace{-2mm}
\subsection{Training Details}
For all experiments on English IWSLT2011, we use the ``BERT-base, Uncased'' model (12 layers, 768 hidden units, 12 heads, 110M parameters) \cite{DBLP:conf/naacl/DevlinCLT19}\footnote{https://github.com/google-research/bert} and ``ELECTRA-large'' model (24 layers, 1024 hidden units, 16 heads, 335M parameters) \cite{DBLP:conf/iclr/ClarkLLM20}\footnote{https://github.com/google-research/electra}.
For IWSLT2011, we train 3 epochs, and use batch size 32 and initial learning rate 5e-5 for BERT-base;  and batch size 16, initial learning rate 2e-5 for ELECTRA-large.
The window size is set to 120, the right overlap size is set to 15, and the left overlap size is set to 35 for BERT-base and 40 for ELECTRA-large.
For the Chinese dataset, we use the ``RoBERTa-wwm-base, Chinese'' models (12 layers, 768 hidden units, 12 heads, 110M parameters) \cite{cui-etal-2020-revisiting}\footnote{https://github.com/ymcui/Chinese-BERT-wwm}.
We use batch size 64, initial learning rate 2e-5, and train 2 epochs. The number of self-training iterations is 1 in all of our experiments.
\subsection{Performance on English IWSLT2011}
\begin{table*}[htb]
\begin{center}
\scalebox{0.9}{
\begin{tabular}{c | c c c | c c c | c c c | c c c}
\hline
\multirow{2}{*}{\textbf{Model}} & 
\multicolumn{3}{c|}{\textbf{Comma}} & \multicolumn{3}{c|}{\textbf{Period}} &
\multicolumn{3}{c|}{\textbf{Question}} & \multicolumn{3}{c}{\textbf{Overall}}\\
& P & R & F$_1$ & P & R & F$_1$ & P & R & F$_1$ & P & R & F$_1$ \\
\hline
T-LSTM \cite{DBLP:conf/interspeech/TilkA15} & 49.6 & 41.4 & 45.1 & 60.2 & 53.4 & 56.6 & 57.1 & 43.5 & 49.4 & 55.0 & 47.2 & 50.8 \\
T-BRNN-pre \cite{DBLP:conf/interspeech/TilkA16} & 65.5 & 47.1 & 54.8 & 73.3 & 72.5 & 72.9 & 70.7 & 63.0 & 66.7 & 70.0 & 59.7 & 64.4 \\
BLSTM-CRF \cite{DBLP:conf/interspeech/YiTWL17} & 58.9 & 59.1 & 59.0 & 68.9 & 72.1 & 70.5 & 71.8 & 60.6 & 65.7 & 66.5 & 63.9 & 65.1 \\
Teacher-Ensemble \cite{DBLP:conf/interspeech/YiTWL17} & 66.2 & 59.9 & 62.9 & 75.1 & 73.7 & 74.4 & 72.3 & 63.8 & 67.8 & 71.2 & 65.8 & 68.4 \\
DRNN-LWMA-pre \cite{DBLP:conf/icassp/Kim19} & 62.9 & 60.8 & 61.9 & 77.3 & 73.7 & 75.5 & 69.6 & 69.6 & 69.6 & 69.9& 67.2 & 68.6  \\
Self-attention-word-speech \cite{DBLP:conf/icassp/YiT19} & 67.4 & 61.1 & 64.1 & 82.5 & 77.4 & 79.9 & 80.1 & 70.2 & 74.8 & 76.7 & 69.6 & 72.9 \\
CT-Transformer \cite{DBLP:conf/icassp/ChenCLW20} & 68.8 & 69.8 & 69.3 & 78.4 & 82.1 & 80.2 & 76.0 & 82.6 & 79.2 & 73.7 & 76.0 & 74.9\\
SAPR \cite{DBLP:conf/icpr/WangCYX18} & 57.2 & 50.8 & 55.9 & 96.7 & 97.3 & 96.8 & 70.6 & 69.2 & 70.3 & 78.2 & 74.4 & 77.4 \\
BERT-base+Adversarial \cite{DBLP:journals/corr/abs-2004-00248} & 76.2 & 71.2 & 73.6 & 87.3 & 81.1 & 84.1 & 79.1 & 72.7 & 75.8 & 80.9 & 75.0 & 77.8 \\
BERT-large+Transfer \cite{DBLP:conf/apsipa/MakhijaHC19} & 70.8 & 74.3 & 72.5 & 84.9 & 83.3 & 84.1 & 82.7 & 93.5 & 87.8 & 79.5 & 83.7 & 81.4 \\
BERT-base+FocalLoss \cite{DBLP:conf/interspeech/YiTTBF20} & 74.4 & 77.1 & 75.7 & 87.9 & 88.2 & 88.1 & 74.2 & 88.5 & 80.7 & 78.8 & 84.6 & 81.6 \\
RoBERTa-large+augmentation \cite{DBLP:conf/aclnut/AlamKA20} & 76.8 & 76.6 & 76.7 & 88.6 & 89.2 & 88.9 & 82.7 & 93.5 & 87.8 & 82.6 & 83.1 & 82.9 \\ 
RoBERTa-base \cite{DBLP:conf/iwslt/CourtlandFM20}  & 76.9 & 75.4 & 76.2 & 86.1 & 89.3 & 87.7 & 88.9 & 87.0 & 87.9 & 84.0 & 83.9 & \textbf{83.9} \\
\hline 
BERT-base & 70.6 & 72.8 & 71.6 & 82.7 & 83.6 & 83.2 & 71.9 & 89.1 & 79.6 & 76.3 & 78.4 & 77.4 \\
BERT-base+Vanilla ST & 73.0 & 74.5 & 73.7 & 82.9 & 85.5 & 84.2 & 74.5 & 89.1 & 81.2 & 77.8 & 80.2 & 79.0 \\
BERT-base+Disc-ST & 73.7 & 74.7 & 74.2 & 83.2 & 86.6 & 84.9 & 75.9 & 89.1 & 82.0 & 78.4 & 80.8 & \textbf{79.6} \\
ELECTRA-large & 76.3 & 81.9 & 79.0 & 89.3 & 90.8 & 90.0 & 79.6 & 93.5 & 86.0 & 82.4 & 86.5 & 84.4 \\
ELECTRA-large+Vanilla ST & 77.4 & 82.0 & 79.6 & 89.5 & 90.5 & 90.0 & 81.1 & 93.5 & 86.9 & 83.1 & 86.4 & 84.7 \\
ELECTRA-large+Disc-ST & 78.0 & 82.4 & 80.1 & 89.9 & 90.8 & 90.4 & 79.6 & 93.5 & 86.0 & 83.6 & 86.7 & \textbf{85.2}  \\
\hline
\end{tabular}
}
\end{center}
\caption{Punctuation prediction results in terms of P(\%), R(\%), F1(\%) on the English IWSLT2011 test set.}
\label{tab:result:en}
\end{table*}

\begin{table*}[htb]
\begin{center}
\scalebox{0.85}{
\begin{tabular}{c |c c c |c c c| c c c |c c c |c c c}
\hline
\multirow{2}{*}{\textbf{Model}} & 
\multicolumn{3}{c|}{\textbf{Comma}} & \multicolumn{3}{c|}{\textbf{Period}} &
\multicolumn{3}{c|}{\textbf{Question}} &
\multicolumn{3}{c|}{\textbf{Enum. Comma}} & \multicolumn{3}{c}{\textbf{Overall}} \\
& P & R & F$_1$ & P & R & F$_1$ & P & R & F$_1$ & P & R & F$_1$ & P & R & F$_1$\\
\hline
BLSTM \cite{DBLP:conf/icassp/ChenCLW20}  & 58.9 & 43.9 & 50.3 & 59.7 & 58.1 & 58.9 & 77.0 & 58.8 & 66.7 & 59.8 & 16.5 & 25.9 & 60.2 & 48.8 & 53.9 \\
Full-Transformer \cite{DBLP:conf/icassp/ChenCLW20}& 61.9 & 50.7 & 55.8 & 60.5 & 64.7 & 62.5 & 74.2 & 68.6 & 71.3 & 64.5 & 30.9 & 41.8 & 62.1 & 55.9 & 58.8 \\
CT-Transformer \cite{DBLP:conf/icassp/ChenCLW20} & 60.8 & 53.5 & 56.9 & 63.8 & 59.7 & 61.7 & 76.3 & 63.0 & 69.0 & 63.4 & 25.2 & 36.1 & 62.7 & 55.3 & 58.8 \\
\hline
RoBERTa-wwm-base & 63.8 & 51.0 & 56.7 & 65.6 & 60.8 & 63.1 & 71.7 & 75.8 & 73.7 & 43.8 & 47.4 & 45.5 & 64.2 & 55.6 & 59.6 \\
RoBERTa-wwm-base + Disc-ST & 62.6 & 53.0 & 57.4 & 64.0 & 64.0 & 64.0 & 73.2 & 71.4 & 72.3 & 49.8 & 34.6 & 40.8 & 63.5 & 57.3 & \textbf{60.2} \\
\hline
\end{tabular}
}
\end{center}
\caption{Punctuation prediction results in terms of P(\%), R(\%), F1(\%) on the internal Chinese test set. }
\label{tab:result:cn}
\end{table*}

We compare the proposed approach with previous models on English IWSLT2011.
The first group of models and results listed in Table~\ref{tab:result:en}  is cited from previous works. T-LSTM \cite{DBLP:conf/interspeech/TilkA15} used uni-directional LSTM and T-BRNN-pre \cite{DBLP:conf/interspeech/TilkA16} used bidirectional RNN with attention. BLSTM-CRF and Teacher-Ensemble are the best single and ensemble models in \cite{DBLP:conf/interspeech/YiTWL17}, respectively.
DRNN-LWMA-pre \cite{DBLP:conf/icassp/Kim19} used a deep recurrent neural network architecture with layer-wise multi-head attentions. 
Self-attention-word-speech \cite{DBLP:conf/icassp/YiT19} used a full sequence Transformer encoder-decoder model with pretrained word2vec and speech2vec embeddings. CT-Transformer \cite{DBLP:conf/icassp/ChenCLW20} used a controllable time-delay Transformer and utilized well-formed text corpora to pretrain. 
SAPR \cite{DBLP:conf/icpr/WangCYX18} used a Transformer encoder-decoder which views the punctuation restoration as a translation task. 
BERT-base+Adversarial \cite{DBLP:journals/corr/abs-2004-00248} used BERT-base with adversarial multi-task learning by adding an extra POS tagging task.
BERT-large+Transfer \cite{DBLP:conf/apsipa/MakhijaHC19} used BERT-large with LSTM-CRF. 
BERT-base+FocalLoss \cite{DBLP:conf/interspeech/YiTTBF20} improved BERT-base with focal loss to alleviate the class imbalance problem. 
RoBERTa-large+augmentation \cite{DBLP:conf/aclnut/AlamKA20} used RoBERTa-large with data augmentation, including insertion, substitution, and deletion. 
The current SOTA (RoBERTa-base) \cite{DBLP:conf/iwslt/CourtlandFM20} used RoBERTa-base with aggregating predictions across multiple context windows.

The second group shows the experimental results from this work (using the decoding strategy in Section~\ref{subsec:decoding}). 
We finetune BERT-base and ELECTRA-large models with the human-labeled training set and observe that ELECTRA-large outperforms the current SOTA (RoBERTa-base) (F$_1$ 84.4 versus 83.9) and significantly outperforms BERT-base (F$_1$ 84.4 versus 77.4). 
Vanilla self-training  (denoted vanilla ST) on BERT-base improves F$_1$ from 77.4 to 79.0 (\textbf{+1.6}), and on ELECTRA-large improves F$_1$ from 84.4 to 84.7 (\textbf{+0.3}), demonstrating that vanilla ST obtains significant gains over the strong baselines.
Replacing vanilla ST with Discriminative Self-Training (Disc-ST) on BERT-base achieves a further improvement on F$_1$ from 79.0 to 79.6 (\textbf{+0.6}), a statistically significant improvement with $p < 0.01$; for ELECTRA-large,  Disc-ST improves F$_1$ from 84.7 by vanilla ST to 85.2 (\textbf{+0.5}), a statistically significant improvement with $p < 0.01$. Our ELECTRA-large+Disc-ST outperforms the current SOTA (RoBERTa-base) by \textbf{1.3} absolute F$_1$ gain.

\subsection{Performance on Chinese Dataset}
We also compare the proposed approach with previous models on the Chinese dataset. The first group of models and results in Table~\ref{tab:result:cn} is cited from previous works. The second group shows the experimental results from this work. 
To compare with previous methods in \cite{DBLP:conf/icassp/ChenCLW20}, we use the same fast decoding strategy \cite{DBLP:conf/icassp/ChenCLW20} with a low frame rate 3 and the number of look-ahead words after end-of-sentence mark as 6.
The RoBERTa-wwm-base model significantly outperforms the current SOTA CT-Transformer (F$_1$ 59.6 versus 58.8). Furthermore, Discriminative Self-Training improves F$_1$ from 59.6 to 60.2 (\textbf{+0.6}), a statistically significant improvement with $p < 0.01$.

\subsection{Ablation Study}
We conduct ablation study to understand the contributions of component algorithms to the overall system performance on the IWSLT2011 test set, as shown in Table \ref{tab:result:ablation}. When we add vanilla ST to BERT-base, F$_1$ improves from 77.4 to 79.0 (\textbf{+1.6}). When we further add weighed loss, F$_1$ improves from 79.0 to 79.3 (\textbf{+0.3}). When we further add equal label smoothing (LS) for both human-labeled data and pseudo-labeled data, F$_1$ improves from 79.3 to 79.4 (\textbf{+0.1}). When we replace equal LS with discriminative LS, F$_1$ improves from 79.3 to 79.6 (\textbf{+0.3}). These results demonstrate that both weighted loss and discriminative LS in discriminative self-training contribute to the improvement over vanilla self-training. 

\begin{table}[htb]
\begin{center}
\begin{tabular}{l  c c c}
\hline
\multirow{2}{*}{\textbf{Model}} & \multicolumn{3}{c}{\textbf{Overall}} \\
& P & R & F$_1$ \\
\hline
BERT-base & 76.3 & 78.4 & 77.4 \\
~+ Vanilla ST & 77.8 & 80.2 & 79.0 \\
~~+ weighted loss & 78.0 & 80.7 & 79.3 \\
~~~+ label smoothing (LS) & 78.4 & 80.4 & 79.4  \\
~~~+ discriminative LS & 78.4 & 80.8 & 79.6\\
\hline
\end{tabular}
\end{center}
\caption{Ablation study of Discriminative Self-training.}
\label{tab:result:ablation}
\end{table}

\vspace{-6mm}
\section{Conclusions}
We  propose a Discriminative Self-Training approach with weighted loss and discriminative label smoothing. Experimental results show that the proposed approach achieves significant improvement on punctuation prediction over strong baselines including BERT, RoBERTa, and ELECTRA models.  The proposed Discriminative Self-Training approach outperforms vanilla self-training. Our approach outperforms the current SOTA on the English IWSLT2011 benchmark and an internal Chinese dataset. Future work includes investigating other output regularizers and the efficacy of self-training in cross-domain and cross-lingual applications.

\bibliographystyle{IEEEtran}
\bibliography{refs,string}

\end{document}